\pdfoutput=1
\documentclass{article} 
\usepackage{graphicx}
\usepackage{iclr2019_conference,times}
\usepackage{color}
\usepackage{subcaption}
\usepackage{soul}
\usepackage{wrapfig}

\usepackage{amsmath,amsfonts,bm}

\def\eqref#1{equation~\ref{#1}}

\def\1{\bm{1}}

\DeclareMathAlphabet{\mathsfit}{\encodingdefault}{\sfdefault}{m}{sl}
\SetMathAlphabet{\mathsfit}{bold}{\encodingdefault}{\sfdefault}{bx}{n}

\usepackage{hyperref}
\usepackage{url}
\usepackage{mathtools}

\title{A closer look at Deep Learning heuristics: Learning rate restarts, warmup and distillation}

\author{Akhilesh Gotmare\thanks{Work performed while interning at Salesforce Research}  \\
Department of Computer Science\\
EPFL, Switzerland \\
\texttt{akhilesh.gotmare@epfl.ch } \\
\And
Nitish Shirish Keskar, Caiming Xiong \& Richard Socher \\
Salesforce Research \\
Palo Alto, US \\
\texttt{\{nkeskar,cxiong,rsocher\}@salesforce.com} \\
}

\iclrfinalcopy 
\begin{document}

\maketitle

\begin{abstract}
The convergence rate and final performance of common deep learning models have significantly benefited from heuristics such as learning rate schedules, knowledge distillation, skip connections, and normalization layers. In the absence of theoretical underpinnings, controlled experiments aimed at explaining these strategies can aid our understanding of deep learning landscapes and the training dynamics. Existing approaches for empirical analysis rely on tools of linear interpolation and visualizations with dimensionality reduction, each with their limitations. Instead, we revisit such analysis of heuristics through the lens of recently proposed methods for loss surface and representation analysis, viz., mode connectivity and canonical correlation analysis (CCA), and hypothesize reasons for the success of the heuristics. In particular, we explore knowledge distillation and learning rate heuristics of (cosine) restarts and warmup using mode connectivity and CCA.  Our empirical analysis suggests that: (a) the reasons often quoted for the success of cosine annealing are not evidenced in practice; (b) that the effect of learning rate warmup is to prevent the deeper layers from creating training instability; and (c) that the latent knowledge shared by the teacher is primarily disbursed to the deeper layers.
\end{abstract}

\section{Introduction}
The introduction of heuristics such as normalization layers \citep{ioffe2015batch, ba2016layer}, residual connections \citep{he2016deep}, and learning rate strategies \citep{loshchilov2016sgdr,goyal2017accurate,smith2017cyclical} have greatly accelerated progress in Deep Learning. Many of these ingredients are now commonplace in modern architectures, and some of them have also been buttressed with theoretical guarantees \citep{balduzzi2017shattered,poggio2017theory,hardt2016identity}. However, despite their simplicity and efficacy, why some of these heuristics work is still relatively unknown. Existing attempts at explaining these strategies empirically have been limited to intuitive explanations and the use of tools such as linear interpolation between two models and low-dimensional visualizations \citep{li2017visualizing} of the loss surface. In our work, we instead use recent tools built specifically for analyzing deep networks, viz., mode connectivity \citep{garipov2018loss} and singular value canonical correlation analysis (SVCCA) \citep{raghu2017svcca}. We investigate three strategies in detail: (a) cosine learning rate decay, (b) learning rate warmup, and (c) knowledge distillation, and list the summary of our contributions at the end of this section. 


Cosine annealing \citep{loshchilov2016sgdr}, also known as stochastic gradient descent with restarts (SGDR), and more generally cyclical learning rate strategies \citep{smith2017cyclical}, have been recently proposed to accelerate training of deep networks \citep{coleman2018analysis}. The strategy involves reductions and restarts of learning rates over the course of training, and was motivated as means to escape spurious local minima. Experimental results have shown that SGDR often improves convergence both from the standpoint of iterations needed for convergence and the final objective. 

Learning rate warmup \citep{goyal2017accurate} also constitutes an important ingredient in training deep networks, especially in the presence of large or dynamic batch sizes. It involves increasing the learning rate to a large value over a certain number of training iterations followed by decreasing the learning rate, which can be performed using step-decay, exponential decay or other such schemes. The strategy was proposed out of the need to induce stability in the initial phase of training with large learning rates (due to large batch sizes). It has been employed in training of several architectures at scale including ResNets and Transformer networks \citep{vaswani2017attention}. 

Further, we investigate knowledge distillation (KD) \citep{hinton2015distilling}. This strategy involves first training a (teacher) model on a typical loss function on the available data. Next, a different (student) model (typically much smaller than the teacher model) is trained, but instead of optimizing the loss function defined using hard data labels, this student model is trained to mimic the teacher model. It has been empirically found that a student network trained in this fashion significantly outperforms an identical network trained with the hard data labels. We defer a detailed discussion of the three heuristics, and existing explanations for their efficacy to sections 3, 4 and 5 respectively.

%

Finally, we briefly describe the tools we employ for analyzing the aforementioned heuristics. Mode connectivity (MC) is a recent observation that shows that, under circumstances, it is possible to connect any two local minima of deep networks via a piecewise-linear curve \citep{garipov2018loss, draxler2018essentially}. This shows that local optima obtained through different means, and exhibiting different local and generalization properties, are connected. The authors propose an algorithm that locates such a curve. While not proposed as such, we employ this framework to better understand loss surfaces but begin our analysis in Section 2 by first establishing its robustness as a framework.

Deep network analyses focusing on the \textit{weights} of a network are inherently limited since there are several invariances in this, such as permutation and scaling. Recently, \citet{raghu2017svcca} propose using CCA along with some pre-processing steps to analyze the \textit{activations} of networks, such that the resulting comparison is not dependent on permutations and scaling of neurons. They also prove the computational gains of using CCA over alternatives (\citep{li2015convergent}) for representational analysis and employ it to better understand many phenomenon in deep learning.

\paragraph{Contributions:}
\begin{itemize}
    \item We use mode connectivity and CCA to improve  understanding of cosine annealing, learning rate warmup and knowledge distillation. For mode connectivity, we also establish the robustness of the approach across changes in training choices for obtaining the modes.
    \item We demonstrate that the reasons often quoted for the success of cosine annealing are not substantiated by our experiments, and that the iterates move over barriers after restarts but the explanation of escaping local minima might be an oversimplification.
    \item We show that learning rate warmup primarily limits weight changes in the deeper layers and that freezing them achieves similar outcomes as warmup.
    \item We show that the latent knowledge shared by the teacher in knowledge distillation is primarily disbursed in the deeper layers.
    
\end{itemize}
\section{Empirical tools}

\subsection{Mode Connectivity} 
\label{mode connectivity}
\citet{garipov2018loss} introduce a framework, called mode connectivity, to obtain a low loss (or high accuracy, in the case of classification) curve of simple form, such as a piecewise linear curve, that connects optima (modes of the loss function) found independently. This observation suggests that points at the same loss function depth are connected, somewhat contrary to several empirical results claiming that minima are isolated or have barriers between them\footnote{\citet{draxler2018essentially} independently report the same observation for neural network loss landscapes, and claim that this is suggestive of the resilience of neural networks to perturbations in model parameters.}. 

Let $w_a \in \mathbb{R}^{D}$ and $w_b \in \mathbb{R}^{D}$ be two modes in the $D$-dimensional parameter space obtained by optimizing a given loss function $\mathcal{L}(w)$ (like the cross-entropy loss). We represent a curve connecting $w_a$ and $w_b$ by 
$\phi_{\theta}(t) : [0,1] \rightarrow \mathbb{R}^{D}$, such that $\phi_{\theta}(0) = w_a$ and $\phi_{\theta}(1) = w_b$. To find a low loss path, we find the set of parameters $\theta \in \mathbb{R}^{D}$ that minimizes the following loss:
$\ell(\theta) = \int_{0}^{1} \mathcal{L}(\phi_{\theta}(t))dt = \mathbb{E}_{t \sim U(0,1)} \mathcal{L}(\phi_{\theta}(t))$
where $U(0,1)$ is the uniform distribution in the interval $[0,1]$.
To optimize $\ell(\theta)$ for $\theta$, we first need to chose a parametric form for $\phi_{\theta}(t)$. One of the forms proposed by \cite{garipov2018loss} is a polygonal chain with a single bend at $\theta$ as follows 
\[   
{\phi_{\theta}(t)} = 
     \begin{cases}
       2 (t {\theta} + (0.5 - t) w_a), &\quad\text{if } 0 \leq t \leq 0.5 \\
       2 ((t - 0.5)w_b + (1 - t){\theta}) &\quad\text{if } 0.5 < t \leq 1 \\
     \end{cases}
\]
To minimize $\ell(\theta)$, we sample $t \sim U[0,1]$ at each iteration and use $\nabla_{\theta}\mathcal{L}(\phi_{\theta}(t))$ as an unbiased estimate for the true gradient $\nabla_{\theta}\ell(\theta)$ to perform updates on $\theta$, where $\theta$ is initialized with $\frac{1}{2}(w_a + w_b)$.

\subsubsection{Resilience of Mode Connectivity}
\label{resilience}

\begin{figure}
    \centering
    \includegraphics[scale = 0.23]{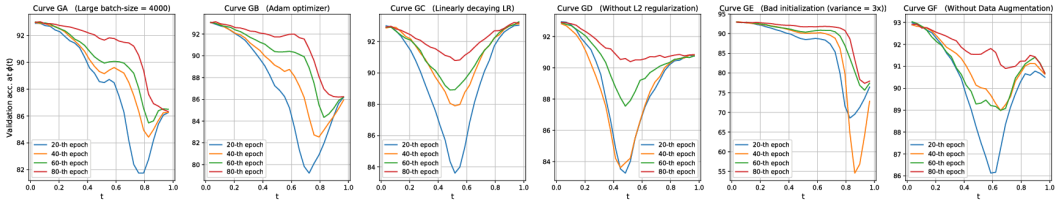}
    \caption{Validation accuracy corresponding to models on the following 6 different curves - curve $GA$ represents curve connecting mode $G$ (one found with default hyperparameters) and mode $A$ (using large batch size), similarly, curve $GB$ connects mode $G$ and mode $B$ (using Adam), curve $GC$ connects to mode $C$ (using linearly decaying learning rate), curve $GD$ to mode $D$ (with lesser L2 regularization), curve $GE$ to mode $E$ (using a poor initialization), and curve $GF$ to mode $F$ (without using data augmentation). $t=0$ corresponds to mode $G$ for all plots.}
    \label{fig:testaccCFsingcol}
\end{figure}

To demonstrate that the curve-finding approach works in practice, \citet{garipov2018loss} use two optima found using different initializations but a common training scheme which we detail below. We explore the limits of this procedure by connecting optima obtained from different training strategies. Our goal of this investigation is to first establish the robustness of the framework in order to seamlessly use it as a tool for analysis. In particular, we experiment with different initializations, optimizers, data augmentation choices, and hyperparameter settings including regularization, training batch sizes, and learning rate schemes. We note in passing that while the framework was proposed to connect two points in the parameter space that are at equal depth in the loss landscape, it is well-defined to also connect points at different depths; in this case, the path corresponds to one that minimizes the average loss along the curve. 

Conventional wisdom suggests that the different training schemes mentioned above will converge to regions in the parameter space that are vastly different from each other. Examples of this include size of minibatches used during training \citep{keskar2016large}, choice of optimizer \citep{heusel2017gans,wilson2017marginal}, initialization \citep{goodfellow2016deep} and choice of regularizer. Having a high accuracy connection between these pairs would seem counterintuitive.


For obtaining the reference model (named mode $G$), we train the VGG-16 model architecture \citep{simonyan2014very} using CIFAR-10 training data \citep{krizhevsky2014cifar} for $200$ epochs with SGD. 
We then build 6 variants of the reference mode $G$ as follows: we obtain mode $A$ using a training batch size of $4000$, mode $B$ by using the Adam optimizer instead of SGD, mode $C$ with a linearly decaying learning rate instead of the step decay used in mode $G$, mode $D$ using a smaller weight decay of $5 \times 10^{-6}$, mode $E$ by increasing the variance of the initialization distribution to $3 \times \sqrt[]{2/n}$ and mode $F$ using no data augmentation. Note that for the set of modes $\{A,B,C,D,E,F\}$, all the other hyper-parameters and settings except the ones mentioned above are kept same as that for mode $G$. We use the mode connectivity algorithm on each of the $6$ pairs of modes including $G$ and another mode, resulting in curves $GA$, $GB$, $GC$, $GD$, $GE$, and $GF$.

Figure \ref{fig:testaccCFsingcol} shows the validation accuracy for models on each of the $6$ connecting curves during the $20$th, $40$th, $60$th and $80$th epochs of the mode connectivity training procedure. As described in Section \ref{mode connectivity}, for a polychain curve $GX$ (connecting modes $G$ and $X$ using the curve described by $\theta$), model parameters $\phi_{\theta}(t)$ on the curve are given by $p_{\phi_{\theta}(t)} = 2 (t p_{\theta} + (0.5 - t) p_G) \; \text{if} \; 0 \leq t \leq 0.5$ and $p_{\phi_{\theta}(t)} = 2 ((t - 0.5)p_X + (1 - t)p_{\theta}) \; \text{if} \; 0.5 < t \leq 1$
where $p_G$, $p_{\theta}$ and $p_X$ are parameters of the models $G$, $\theta$, and $X$ respectively. Thus $\phi_{\theta}(0) = G$ and $\phi_{\theta}(1) = X$.

In a few epochs of the curve training, for all 6 pairs, we can find a curve such that each point on it generalizes almost as well as models from the pair that is being connected. Note that by virtue of existence of these $6$ curves, there exists a high accuracy connecting curve (albeit with multiple bends) for each of the ${7\choose 2}$ pairs of modes. We refer the reader to Appendix~\ref{sec:appendixmc} for a t-SNE plot of the modes and their connections, and also for additional plots and details. Having established the high likelihood of the existence of these curves, we use this procedure along with interpolation of the loss surface between parameters at different epochs as tools to analyze the dynamics of SGD and SGDR.

\subsection{CCA for measuring representational similarity}
\label{cca}
Canonical correlation analysis (CCA) is a classical tool from multivariate statistics \citep{hotelling1936relations} that investigates the relationships between two sets of random variables. \cite{raghu2017svcca} have proposed coupling CCA with pre-processing steps like Singular Value Decomposition (SVD) or Discrete Fourier Transform (DFT) to design a similarity metric for two neural net layers that we want to compare. These layers do not have to be of the same size or belong to the same network. 

Given a dataset with $m$ examples $X = \{x_1, \dots x_m \}$, we denote the scalar output of the neuron $z_i^{l}$ ($i$-th neuron of layer $l$) for the input $x_i$ by $f_{z_{i}^L}(x_i)$. These scalar outputs can be stacked (along $n$ different neurons and $m$ different datapoints) to create a matrix $L\in \mathbb{R}^{m \times n}$ representing the output of a layer corresponding to the entire dataset. This choice of comparing neural network layers using activations instead of weights and biases is crucial to the setup proposed. Indeed, invariances due to re-parameterizations and permutations limit the interpretability of the model weights \citep{dinh}. However, under CCA of the layers, two activation sets are comparable by design.

Given representations corresponding to two layers $L_a \in \mathbb{R}^{m_a \times n}$ and $L_b \in \mathbb{R}^{m_b \times n}$, SVCCA first performs dimensionality reduction using SVD to obtain $L_a^{'} \in \mathbb{R}^{m_a' \times n}$ and $L_{b}^{'} \in \mathbb{R}^{m_b' \times n}$ while preserving $99\%$ of the variance. The subsequent CCA step involves transforming $L_a^{'}$ and $L_{b}^{'}$ to $a_{1}^{\top}L_a^{'}$ and $b_{1}^{\top}L_b^{'}$ respectively where $\{a_{1},b_{1}\}$ is found by maximizing the correlation between the transformed subspaces, 
and the corresponding correlation is denoted by $\rho_1$. This process continues, using orthogonality constraints, till $c = \text{min}\{m_a^{'},m_b^{'}\}$ leading to the set of correlation values $\{ \rho_{1}, \rho_{2} \dots \rho_{c} \}$ corresponding to $c$ pairs of canonical variables $\{ \{a_1,b_1\},\{a_2,b_2\}, \dots \{a_c,b_c\}\}$ respectively. We refer the reader to \cite{raghu2017svcca} for details on solving these optimization problems. The average of these $c$ correlations $\frac{1}{n}\sum_{i}\rho_i$ is then considered as a measure of the similarity between the two layers. For convolutional layers, \cite{raghu2017svcca} suggest using a DFT pre-processing step before CCA, since they typically have a large number of neurons ($m_a$ or $m_b$), where performing raw SVD and CCA would be computationally too expensive. This procedure can then be employed to compare different neural network representations and to determine how representations evolve over training iterations.
%




\section{Stochastic Gradient Descent with Restarts (SGDR)}
\label{sgdwithwarmrestarts}

\begin{figure}
    \centering
    \includegraphics[ width = \textwidth]{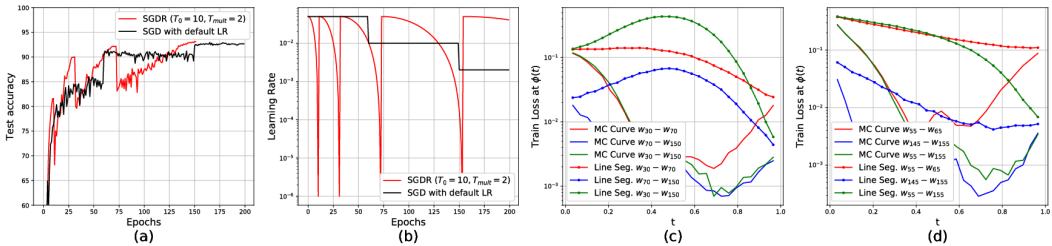}
    \caption{(a) Validation accuracy of a VGG16 model trained on CIFAR-10 using SGDR with warm restarts simulated every $T_0 = 10$ epochs and $T_{mult} = 2$. (b) SGDR and SGD learning rate schemes. (c) Cross-entropy training loss on the curve found through Mode Connectivity (MC Curve) and on the line segment (Line Seg.) joining modes $w_{30}$ (model corresponding to parameters at the $30$-th epoch of \textbf{SGDR}) and $w_{70}$, $w_{70}$ and $w_{150}$, $w_{30}$ and $w_{150}$. (d) Cross-entropy training loss on the curve found through Mode Connectivity (MC Curve) and on the line segment (Line Seg.) joining modes $w_{55}$ (model corresponding to parameters at the $55$-th epoch of \textbf{SGD with step decay learning rate scheme}) and $w_{65}$, $w_{145}$ and $w_{155}$, $w_{55}$ and $w_{155}$.}
    \label{fig:sgdr_sgd_modeconn}
\end{figure}

\citet{loshchilov2016sgdr} introduced SGDR as a modification to the common linear or step-wise decay of learning rates. The strategy decays learning rates along a cosine curve and then, at the end of the decay, restarts them to its initial value. 
The learning rate at the $t$-th epoch in SGDR is given by the following expression in (\ref{sgdreqn}) where $\eta_{min}$ and $\eta_{max}$ are the lower and upper bounds respectively for the learning rate. $T_{cur}$ represents how many epochs have been performed since the last restart and a warm restart is simulated once $T_i$ epochs are performed. Also $T_i = T_{mult} \times T_{i-1}$, meaning the period $T_{i}$ for the learning rate variation is increased by a factor of $T_{mult}$ after each restart. Figure \ref{fig:sgdr_sgd_modeconn}(b) shows an instance of this learning rate schedule.
\begin{equation}
\label{sgdreqn}
\eta_{t} = \eta_{min} + \frac{1}{2}(\eta_{max} - \eta_{min})\left(1 +\cos \left(\frac{T_{cur}}{T_i}\pi\right)\right)
\end{equation}
While the strategy has been claimed to outperform other learning rate schedulers, little is known why this has been the case. One explanation that has been given in support of SGDR is that it can be useful to deal with multi-modal functions, where the iterates could get stuck in a local optimum and a restart will help them get out of it and explore another region; however, \citet{loshchilov2016sgdr} do not claim to observe any effect related to multi-modality. \citet{huang2017snapshot} propose an ensembling strategy using the set of iterates before restarts and claim that, when using the learning rate annealing cycles, the optimization path converges to and escapes from several local minima. We empirically investigate if this is actually the case by interpolating the loss surface between parameters at different epochs and studying the training and validation loss for parameters on the hyperplane passing through\footnote{This hyperplane is the set of all affine combinations of $w_a, w_b \text{   and their connection } \theta = w_{a-b}$.} the two modes found by SGDR and their connectivity. Further, by employing the CCA framework as described in Section \ref{cca}, we investigate the progression of training, and the effect of restarts on the model activations.

We train a VGG-16 network \citep{simonyan2014very} on the CIFAR-10 dataset using SGDR. For our experiments, we choose $T_0 = 10$ epochs and $T_{mult} = 2$ (warm restarts simulated every 10 epochs and the period $T_i$ doubled at every new warm restart), $\eta_{max} = 0.05$ and $\eta_{min} = 10^{-6}$. We also perform VGG training using SGD (with momentum of $0.9$) and a step decay learning rate scheme (initial learning rate of $\eta_{0} = 0.05$, scaled by $5$ at epochs $60$ and $150$). 
Figure \ref{fig:sgdr_sgd_modeconn}(b) shows the learning rate variation for these two schemes on a logarithmic scale and Figure \ref{fig:sgdr_sgd_modeconn}(a) shows the validation accuracy over training epochs with these two learning rate schemes. 

\begin{wrapfigure}{r}{0.5\textwidth}
    \centering
    \includegraphics[ width = 0.48\textwidth]{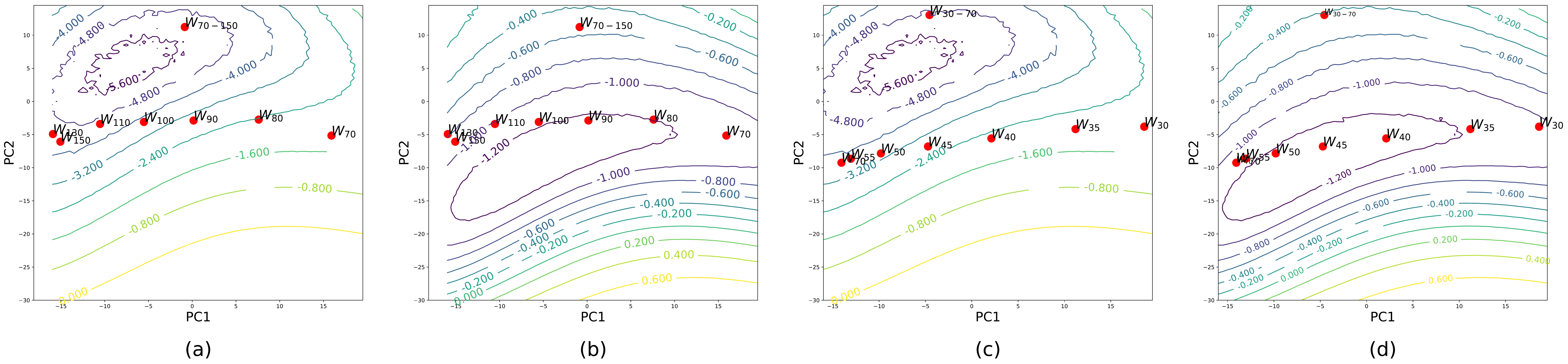}
    \caption{(a) Training loss surface and (b) validation loss surface, log scales, for points on the plane defined by $\{w_{70},w_{150},w_{70-150}\}$ including projections of the SGDR iterates on this hyperplane.}
    \label{fig:sgdr_mc_contours}
    \vspace{-0.25cm}
\end{wrapfigure}

In order to understand the loss landscape on the optimization path of SGDR, the pairs of iterates obtained just before the restarts 
$\{w_{30},w_{70}\},\{w_{70},w_{150}\}$ and $\{w_{30},w_{150}\}$ are given as inputs to the mode connectivity algorithm, where $w_n$ is the model corresponding to parameters at the $n$-th epoch of training. Figure \ref{fig:sgdr_sgd_modeconn}(c) shows the training loss for models along the line segment joining these pairs and those on the curve found through mode connectivity. For the baseline case of SGD training, we connect the iterates around the epochs when we decrease our learning rate in the step decay learning rate scheme. Thus, we chose $\{w_{55},w_{65}\},\{w_{145},w_{165}\}$ and $\{w_{55},w_{165}\}$ as input pairs to the mode connectivity algorithm.  Figure \ref{fig:sgdr_sgd_modeconn}(d) shows the training loss for models along the line segments joining these pairs and the curves found through mode connectivity.

From Figure \ref{fig:sgdr_sgd_modeconn}(c), it is clear that for the pairs $\{w_{30},w_{150}\}$ and $\{w_{70},w_{150}\}$ the training loss for points on segment is much higher than the endpoints suggesting that SGDR indeed finds paths that move over a barrier\footnote{a path is said to have moved over or crossed a barrier between epoch $m$ and $n$ ($n > m$) if $\exists$ $w_t \in \{\lambda w_m + (1 - \lambda) w_n | \lambda \in  [0,1] \}$ such that $\mathcal{L}(w_t) > \max \{\mathcal{L}(w_m),\mathcal{L}(w_n) \} $} in the training loss landscape. In contrast, for SGD (without restarts) in Figure \ref{fig:sgdr_sgd_modeconn}(d) none of the three pairs show evidence of having a training loss barrier on the line segment joining them. Instead there seems to be an almost linear decrease of training loss along the direction of these line segments, suggesting that SGD's trajectory is quite different from SGDR's. We present additional experiments, including results for other metrics, in Appendix~\ref{sec:appendixsgdr}.

To further understand the SGDR trajectory, we evaluate the intermediate points on the hyperplane in the $D$-dimensional space defined by the three points: $w_{70}$, $w_{150}$ and $w_{70-150}$, where $w_{70-150}$ is the bend point that defines the high accuracy connection for the pair $\{w_{70},w_{150}\}$. Figures \ref{fig:sgdr_mc_contours}(a) and \ref{fig:sgdr_mc_contours}(b) show the training and validation loss surface for points in this subspace, respectively. Note that the intermediate iterates do not necessarily lie in this plane, and thus are projected. We refer the reader to Appendix~\ref{sec:appendixsgdr} for additional details on the projection, and analogous results with $w_{30}$ and $w_{70}$. 

Figure \ref{fig:sgdr_mc_contours}(a) suggests that SGDR helps the iterates converge to a different region although neither of $w_{70}$ or $w_{150}$ are technically a local minimum, nor do they appear to be lying in different \textit{basins}, hinting that \citet{huang2017snapshot}'s claims about SGDR converging to and escaping from local minima might be an oversimplification.\footnote{We note in passing that during our experiments, a strategy that consistently performed well is one of a cosine (or linear) decay over the entire budget. We hypothesize that this decay, and less so the restarts, plays a major role in the success of such strategies. } 
Another insight we can draw from Figure \ref{fig:sgdr_mc_contours}(a) is that the path found by mode connectivity corresponds to lower training loss than the loss at the iterates that SGDR converges to ($\mathcal{L}(w_{150}) > \mathcal{L}(w_{70-150})$). However, Figure \ref{fig:sgdr_mc_contours}(b) shows that models on this curve seem to overfit and not generalize as well as the iterates $w_{70}$ and $w_{150}$. Thus, although gathering models from this connecting curve might seem as a novel and computationally cheap way of creating ensembles, this generalization gap alludes to one limitation in doing so; \citet{garipov2018loss} point to other shortcomings of curve ensembling in their original work. 
In Figure \ref{fig:sgdr_mc_contours}, the region of the plane between the iterates $w_{70}$ and $w_{150}$ corresponds to higher training loss but lower validation loss than the two iterates. This hints at a reason why averaging iterates to improve generalization using cyclic or constant learning rates \citep{izmailov2018averaging} has been found to work well.   

Finally, in Figure \ref{fig:SGDR_heatmaps} in Appendix~\ref{sec:appendixsgdrsvcca}, we present the CCA similarity plots for two pairs of models: epochs 10 and 150 (model at the beginning and end of training), and epochs 150 and 155 (model just before and just after a restart). For standard SGD training, \cite{raghu2017svcca} observe that the activations of the shallower layers bear closer resemblance than the deeper layers between a partially and fully trained network from a given training run. For SGDR training, we witness similar results (discussed in Appendix~\ref{sec:appendixsgdrsvcca}), meaning that the representational similarities between the network layers at the beginning and end of training are alike for SGDR and SGD, even though restarts lead to a trajectory that tends to cross over barriers. 

\section{Warmup learning rate scheme}

Learning rate warmup is a common heuristic used by many practitioners for training deep neural nets for computer vision \citep{goyal2017accurate} and natural language processing \citep{bogoychev2018accelerating, vaswani2017attention} tasks. 
Theoretically, it can be shown that the learning dynamics of SGD rely on the ratio of the batch size and learning rate \citep{smith2017don,jastrzkebski2017three,hoffer2017train}. And hence, an increase in batch size over a baseline requires an accompanying increase in learning rate for comparable training. However, in cases when the batch size is increased significantly, the curvature of the loss function typically does not support a proportional increase in the learning rate. Warmup is hence motivated as a means to use large learning rates without causing training instability.
We particularly focus on the importance of the learning rate schedule's warmup phase in the large batch (LB) training of deep convolutional neural networks as discussed in \cite{goyal2017accurate}. Their work adopts a linear scaling rule for adjusting the learning rate as a function of the minibatch size, to enable large-batch training. The question we aim to investigate here is: \textit{How does learning rate warmup impact different layers of the network?}
\begin{figure}
    \centering
    \includegraphics[width=\textwidth]{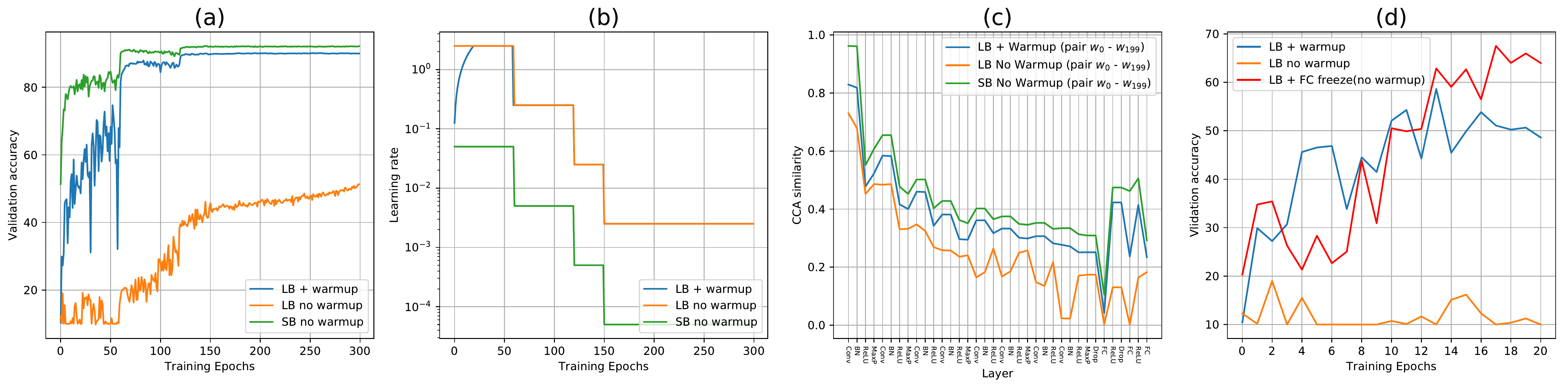}
    \caption{(a) Validation accuracy and (b) Learning rate for the three training setups (c) CCA similarity for $i$-th layer from two different iterations ($0$-th (before warmup) and $200$-th (after warmup) during training (d) Comparing warmup and FC freezing strategies on VGG11 training}
    \label{fig:warmup_lr}
\end{figure}

Using CCA as a tool to study the learning dynamics of neural networks through training iterations, we investigate the differences and similarities for the following $3$ training configurations - (a) large batch training with warmup (LB + warmup), (b) large batch training without warmup (LB no warmup) and (c) small batch training without warmup (SB no warmup). 
We train a VGG-11 architecture on the CIFAR-10 \citep{krizhevsky2014cifar} dataset using SGD with momentum of $0.9$. Learning rate for the small batch case (batch-size of $100$) is set to $0.05$, and for the large batch cases (batch-size of $5000$) is set to $2.5$ as per the scaling rule. For the warmup, we increase the learning rate from $0$ to $2.5$ over the first $200$ iterations. Subsequently, we decrease the learning rate as per the step decay schedule for all runs, scaling it down by a factor of $10$ at epochs $60$, $120$ and $150$. We plot the learning rate and validation accuracy for these $3$ cases in Figure \ref{fig:warmup_lr}(b) and (a).

\begin{figure}
    \centering
    \includegraphics[width=\textwidth]{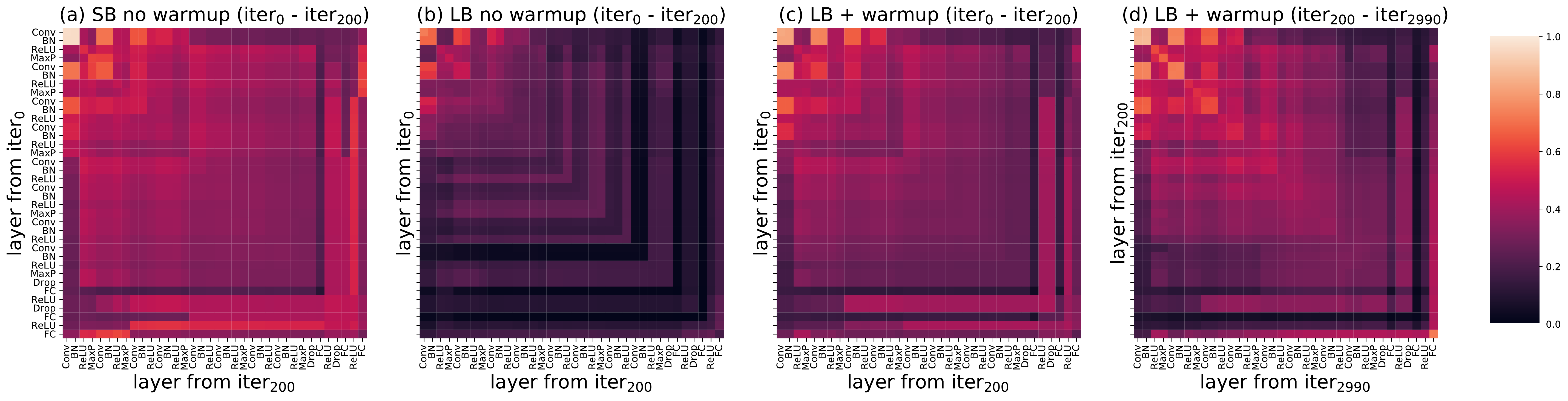}
    \caption{CCA similarity output plots for (\textbf{a}) SB no warmup, (\textbf{b}) LB no warmup, (\textbf{c, d}) LB + warmup training. The $i,j$-th cell represents the CCA similarity between layer $i$ of the first model, and layer $j$ of other. A higher score implies that the layers are more similar (lighter color).}
    \label{fig:warmup_heatmaps}
\end{figure}

Using CCA and denoting the model at the $j$-th iteration of a training setup by $\textit{iter}_j$, we compare activation layers from $\textit{iter}_0$ (init.) and $\textit{iter}_{200}$ (end of warmup) for each of the three runs, presented in Figures \ref{fig:warmup_heatmaps}(a), (b) and (c), and also layers from $\textit{iter}_{200}$ (end of warmup) and $\textit{iter}_{2990}$ (end of training) for the LB + warmup case, presented in Figure \ref{fig:warmup_heatmaps}(d). Figure \ref{fig:warmup_lr}(c) plots the similarity for layer $i$ of $\textit{iter}_a$ with the same layer of $\textit{iter}_b$ (this corresponds to diagonal elements of the matrices in Figure \ref{fig:warmup_heatmaps}) for these three setups. 

An evident pattern in Figures \ref{fig:warmup_heatmaps}(a), (b) and (c) is the increase in similarity for the last few layers (stack of fully-connected layers) for the LB + warmup and SB cases, which is absent in the LB without warmup case. This suggests that when used with the large batch size and learning rate, warmup tends to avoid unstably large changes in the fully-connected (FC) stack for this network configuration. To validate this proposition, we train using the LB without warmup setup, but freezing the fully-connected stack for the first $200$ iterations (LB no warmup + FC freeze). Figure \ref{fig:warmup_lr}(d) shows the validation accuracy for this training run in comparison to the three training setups discussed before. The performance is comparable at the end of warmup by freezing the FC stack, suggesting the validity our proposition in this case. 
We refer the reader to Appendix~\ref{sec:appendixresnetwarmups} for analogous results for ResNet-18 and ResNet-32 \citep{he2016deep}; thus also demonstrating the generality of our claim.
Finally, note from Figure~\ref{fig:warmup_lr}(d) that no qualitative difference exists in the trajectory beyond the warmup when compared to the standard training approach \citep{raghu2017svcca}.
\section{Knowledge Distillation}

We study knowledge distillation as proposed by \cite{hinton2015distilling} using CCA to measure representational similarity between layers of the teacher and student model. Distillation involves training a ``student'' model using the output probability distribution of a ``teacher'' model. This has been widely known to help the student model perform better than it would, if it were trained using hard labels due to knowledge transfer from the teacher model. The reason often quoted for the success of distillation is the transfer of \textit{dark knowledge} from the teacher to the student \citep{hinton2015distilling}, and more recently, as an interpretation of importance weighing \citep{furlanello2018born}. We investigate if this knowledge transfer is limited to certain parts of the network, and if representational similarity between layers of the student and teacher model and a student can help answer this question. 

To construct an example of distillation that can be used for our analysis, we use a VGG-16 model \citep{simonyan2014very} as our teacher network and a shallow convolutional network (\texttt{[conv, maxpool, relu]x2, fc, relu, fc, fc, softmax}) as the student network. We train the shallow network for CIFAR-10 using the teacher's predicted probability distribution (softened using a temperature of $5$), ($S_{\text{distilled}}$), and for the baseline, train another instance of the same model in a standard way using hard labels, ($S_{\text{indep.}}$). 
Over 5 runs for each of the two setups, we find the distillation training attains the best validation accuracy at $85.18\%$ while standard training attains its best at $83.01\%$. We compare their layer-wise representations with those of the teacher network ($T$). 

\begin{figure}
    \centering
    \includegraphics[width=\textwidth]{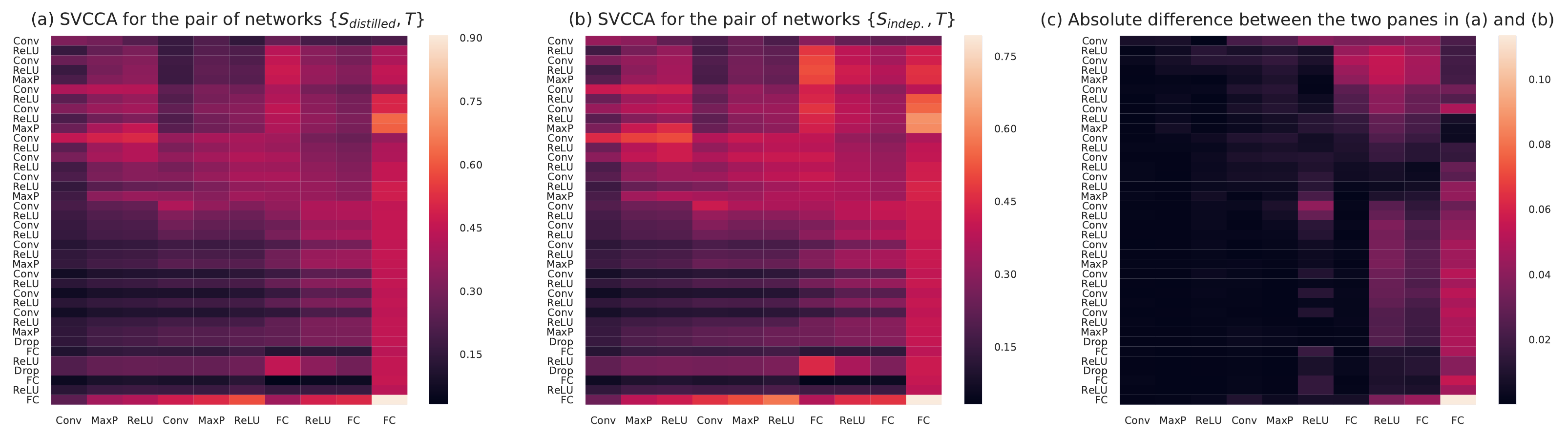}
    \caption{CCA similarity between $S_{\text{distilled}}$ - $T$, $S_{\text{indep.}}$ - $T$, and their difference. $i,j$-th cell of the difference plot represents $ | \text{CCA}(l^{i}_T , l^{j}_{S_{\text{distilled}}}) - \text{CCA}(l^{i}_T ,l^{j}_{S_{\text{indep.}}} ) |$ where $l^{i}_{M}$ denotes the $i$-th layer of network $M$, $T$ denotes the teacher network (VGG16), $S_{\text{distilled}}$ is the student network trained using distillation and $S_{\text{indep.}}$ is the student network trained using hard training labels.}

    \label{fig:distill}
\end{figure}

Figure \ref{fig:distill} shows the CCA plots and the absolute value of their difference. The scores of these two pairs are quite similar for the shallow layers of the student network relative to the deeper layers, suggesting that the difference that knowledge distillation brings to the training of smaller networks is restricted to the deeper layers (\texttt{fc} stack). Similar results are obtained through different configurations for the student and teacher when the student benefits from the teacher's knowledge. We hypothesize that the \textit{dark} knowledge transferred by the teacher is localized majorly in the deeper (discriminative) layers, and less so in the feature extraction layers. We also note that this is not dissimilar to the hypothesis of \citet{furlanello2018born}, and also relates ot the results from the literature on fine-tuning  or transfer learning \citep{goodfellow2016deep,yosinski2014transferable,howard2018universal} which suggest training of only higher layers. 

\section{Conclusion}
Heuristics have played an important role in accelerating progress of deep learning. Founded in empirical experience, intuition and observations, many of these strategies are now commonplace in architectures. In the absence of strong theoretical guarantees, controlled experiments aimed at explaining the the efficacy of these strategies can aid our understanding of deep learning and the training dynamics. We investigate three such heuristics: cosine annealing, learning rate warmup, and knowledge distillation. For this purpose, we employ recently proposed tools of mode connectivity and CCA. Our empirical analysis sheds light on these heuristics and suggests that: (a) the reasons often quoted for the success of cosine annealing are not evidenced in practice; (b) that the effect of learning rate warmup is to prevent the deeper layers from creating training instability; and (c) that the latent knowledge shared by the teacher is primarily disbursed in the deeper layers. Our empirical experiments and hypotheses open new questions, and encourage a deeper exploration into improving and better understanding these heuristics.

\bibliography{iclr2019_conference}

\begin{thebibliography}{35}
\providecommand{\natexlab}[1]{#1}
\providecommand{\url}[1]{\texttt{#1}}
\expandafter\ifx\csname urlstyle\endcsname\relax
  \providecommand{\doi}[1]{doi: #1}\else
  \providecommand{\doi}{doi: \begingroup \urlstyle{rm}\Url}\fi

\bibitem[Ba et~al.(2016)Ba, Kiros, and Hinton]{ba2016layer}
Jimmy~Lei Ba, Jamie~Ryan Kiros, and Geoffrey~E Hinton.
\newblock Layer normalization.
\newblock \emph{arXiv preprint arXiv:1607.06450}, 2016.

\bibitem[Balduzzi et~al.(2017)Balduzzi, Frean, Leary, Lewis, Ma, and
  McWilliams]{balduzzi2017shattered}
David Balduzzi, Marcus Frean, Lennox Leary, JP~Lewis, Kurt Wan-Duo Ma, and
  Brian McWilliams.
\newblock The shattered gradients problem: If resnets are the answer, then what
  is the question?
\newblock \emph{arXiv preprint arXiv:1702.08591}, 2017.

\bibitem[Bogoychev et~al.(2018)Bogoychev, Junczys-Dowmunt, Heafield, and
  Aji]{bogoychev2018accelerating}
Nikolay Bogoychev, Marcin Junczys-Dowmunt, Kenneth Heafield, and Alham~Fikri
  Aji.
\newblock Accelerating asynchronous stochastic gradient descent for neural
  machine translation.
\newblock \emph{arXiv preprint arXiv:1808.08859}, 2018.

\bibitem[Coleman et~al.(2018)Coleman, Kang, Narayanan, Nardi, Zhao, Zhang,
  Bailis, Olukotun, Re, and Zaharia]{coleman2018analysis}
Cody Coleman, Daniel Kang, Deepak Narayanan, Luigi Nardi, Tian Zhao, Jian
  Zhang, Peter Bailis, Kunle Olukotun, Chris Re, and Matei Zaharia.
\newblock Analysis of dawnbench, a time-to-accuracy machine learning
  performance benchmark.
\newblock \emph{arXiv preprint arXiv:1806.01427}, 2018.

\bibitem[Dinh et~al.(2017)Dinh, Pascanu, Bengio, and Bengio]{dinh}
Laurent Dinh, Razvan Pascanu, Samy Bengio, and Yoshua Bengio.
\newblock Sharp minima can generalize for deep nets.
\newblock \emph{arXiv preprint arXiv:1703.04933}, 2017.

\bibitem[Draxler et~al.(2018)Draxler, Veschgini, Salmhofer, and
  Hamprecht]{draxler2018essentially}
Felix Draxler, Kambis Veschgini, Manfred Salmhofer, and Fred~A Hamprecht.
\newblock Essentially no barriers in neural network energy landscape.
\newblock \emph{arXiv preprint arXiv:1803.00885}, 2018.

\bibitem[Furlanello et~al.(2018)Furlanello, Lipton, Tschannen, Itti, and
  Anandkumar]{furlanello2018born}
Tommaso Furlanello, Zachary~C Lipton, Michael Tschannen, Laurent Itti, and
  Anima Anandkumar.
\newblock Born again neural networks.
\newblock \emph{arXiv preprint arXiv:1805.04770}, 2018.

\bibitem[Garipov et~al.(2018)Garipov, Izmailov, Podoprikhin, Vetrov, and
  Wilson]{garipov2018loss}
Timur Garipov, Pavel Izmailov, Dmitrii Podoprikhin, Dmitry~P Vetrov, and
  Andrew~Gordon Wilson.
\newblock Loss surfaces, mode connectivity, and fast ensembling of dnns.
\newblock \emph{arXiv preprint arXiv:1802.10026}, 2018.

\bibitem[Goodfellow et~al.(2016)Goodfellow, Bengio, Courville, and
  Bengio]{goodfellow2016deep}
Ian Goodfellow, Yoshua Bengio, Aaron Courville, and Yoshua Bengio.
\newblock \emph{Deep learning}, volume~1.
\newblock MIT Press, 2016.

\bibitem[Goyal et~al.(2017)Goyal, Doll{\'a}r, Girshick, Noordhuis, Wesolowski,
  Kyrola, Tulloch, Jia, and He]{goyal2017accurate}
Priya Goyal, Piotr Doll{\'a}r, Ross Girshick, Pieter Noordhuis, Lukasz
  Wesolowski, Aapo Kyrola, Andrew Tulloch, Yangqing Jia, and Kaiming He.
\newblock Accurate, large minibatch sgd: training imagenet in 1 hour.
\newblock \emph{arXiv preprint arXiv:1706.02677}, 2017.

\bibitem[Hardt \& Ma(2016)Hardt and Ma]{hardt2016identity}
Moritz Hardt and Tengyu Ma.
\newblock Identity matters in deep learning.
\newblock \emph{arXiv preprint arXiv:1611.04231}, 2016.

\bibitem[He et~al.(2016)He, Zhang, Ren, and Sun]{he2016deep}
Kaiming He, Xiangyu Zhang, Shaoqing Ren, and Jian Sun.
\newblock Deep residual learning for image recognition.
\newblock In \emph{Proceedings of the IEEE conference on computer vision and
  pattern recognition}, pp.\  770--778, 2016.

\bibitem[Heusel et~al.(2017)Heusel, Ramsauer, Unterthiner, Nessler, Klambauer,
  and Hochreiter]{heusel2017gans}
Martin Heusel, Hubert Ramsauer, Thomas Unterthiner, Bernhard Nessler,
  G{\"u}nter Klambauer, and Sepp Hochreiter.
\newblock Gans trained by a two time-scale update rule converge to a nash
  equilibrium.
\newblock \emph{arXiv preprint arXiv:1706.08500}, 2017.

\bibitem[Hinton et~al.(2015)Hinton, Vinyals, and Dean]{hinton2015distilling}
Geoffrey Hinton, Oriol Vinyals, and Jeff Dean.
\newblock Distilling the knowledge in a neural network.
\newblock \emph{arXiv preprint arXiv:1503.02531}, 2015.

\bibitem[Hoffer et~al.(2017)Hoffer, Hubara, and Soudry]{hoffer2017train}
Elad Hoffer, Itay Hubara, and Daniel Soudry.
\newblock Train longer, generalize better: closing the generalization gap in
  large batch training of neural networks.
\newblock In \emph{Advances in Neural Information Processing Systems}, pp.\
  1731--1741, 2017.

\bibitem[Hotelling(1936)]{hotelling1936relations}
Harold Hotelling.
\newblock Relations between two sets of variates.
\newblock \emph{Biometrika}, 28\penalty0 (3/4):\penalty0 321--377, 1936.

\bibitem[Howard \& Ruder(2018)Howard and Ruder]{howard2018universal}
Jeremy Howard and Sebastian Ruder.
\newblock Universal language model fine-tuning for text classification.
\newblock In \emph{Proceedings of the 56th Annual Meeting of the Association
  for Computational Linguistics (Volume 1: Long Papers)}, volume~1, pp.\
  328--339, 2018.

\bibitem[Huang et~al.(2017)Huang, Li, Pleiss, Liu, Hopcroft, and
  Weinberger]{huang2017snapshot}
Gao Huang, Yixuan Li, Geoff Pleiss, Zhuang Liu, John~E Hopcroft, and Kilian~Q
  Weinberger.
\newblock Snapshot ensembles: Train 1, get m for free.
\newblock \emph{arXiv preprint arXiv:1704.00109}, 2017.

\bibitem[Ioffe \& Szegedy(2015)Ioffe and Szegedy]{ioffe2015batch}
Sergey Ioffe and Christian Szegedy.
\newblock Batch normalization: Accelerating deep network training by reducing
  internal covariate shift.
\newblock \emph{arXiv preprint arXiv:1502.03167}, 2015.

\bibitem[Izmailov et~al.(2018)Izmailov, Podoprikhin, Garipov, Vetrov, and
  Wilson]{izmailov2018averaging}
Pavel Izmailov, Dmitrii Podoprikhin, Timur Garipov, Dmitry Vetrov, and
  Andrew~Gordon Wilson.
\newblock Averaging weights leads to wider optima and better generalization.
\newblock \emph{arXiv preprint arXiv:1803.05407}, 2018.

\bibitem[Jastrzebski et~al.(2017)Jastrzebski, Kenton, Arpit, Ballas, Fischer,
  Bengio, and Storkey]{jastrzkebski2017three}
Stanis{\l}aw Jastrzebski, Zachary Kenton, Devansh Arpit, Nicolas Ballas, Asja
  Fischer, Yoshua Bengio, and Amos Storkey.
\newblock Three factors influencing minima in sgd.
\newblock \emph{arXiv preprint arXiv:1711.04623}, 2017.

\bibitem[Keskar et~al.(2016)Keskar, Mudigere, Nocedal, Smelyanskiy, and
  Tang]{keskar2016large}
Nitish~Shirish Keskar, Dheevatsa Mudigere, Jorge Nocedal, Mikhail Smelyanskiy,
  and Ping Tak~Peter Tang.
\newblock On large-batch training for deep learning: Generalization gap and
  sharp minima.
\newblock \emph{arXiv preprint arXiv:1609.04836}, 2016.

\bibitem[Krizhevsky et~al.(2014)Krizhevsky, Nair, and
  Hinton]{krizhevsky2014cifar}
Alex Krizhevsky, Vinod Nair, and Geoffrey Hinton.
\newblock The cifar-10 dataset.
\newblock \emph{online: http://www. cs. toronto. edu/kriz/cifar. html}, 2014.

\bibitem[Li et~al.(2017)Li, Xu, Taylor, and Goldstein]{li2017visualizing}
Hao Li, Zheng Xu, Gavin Taylor, and Tom Goldstein.
\newblock Visualizing the loss landscape of neural nets.
\newblock \emph{arXiv preprint arXiv:1712.09913}, 2017.

\bibitem[Li et~al.(2015)Li, Yosinski, Clune, Lipson, and
  Hopcroft]{li2015convergent}
Yixuan Li, Jason Yosinski, Jeff Clune, Hod Lipson, and John~E Hopcroft.
\newblock Convergent learning: Do different neural networks learn the same
  representations?
\newblock In \emph{FE@ NIPS}, pp.\  196--212, 2015.

\bibitem[Loshchilov \& Hutter(2016)Loshchilov and Hutter]{loshchilov2016sgdr}
Ilya Loshchilov and Frank Hutter.
\newblock Sgdr: stochastic gradient descent with restarts.
\newblock \emph{arXiv preprint arXiv:1608.03983}, 2016.

\bibitem[Maaten \& Hinton(2008)Maaten and Hinton]{maaten2008visualizing}
Laurens van~der Maaten and Geoffrey Hinton.
\newblock Visualizing data using t-sne.
\newblock \emph{Journal of machine learning research}, 9\penalty0
  (Nov):\penalty0 2579--2605, 2008.

\bibitem[Poggio \& Liao(2017)Poggio and Liao]{poggio2017theory}
Tomaso Poggio and Qianli Liao.
\newblock \emph{Theory ii: Landscape of the empirical risk in deep learning}.
\newblock PhD thesis, Center for Brains, Minds and Machines (CBMM), arXiv,
  2017.

\bibitem[Raghu et~al.(2017)Raghu, Gilmer, Yosinski, and
  Sohl-Dickstein]{raghu2017svcca}
Maithra Raghu, Justin Gilmer, Jason Yosinski, and Jascha Sohl-Dickstein.
\newblock Svcca: Singular vector canonical correlation analysis for deep
  learning dynamics and interpretability.
\newblock In \emph{Advances in Neural Information Processing Systems}, pp.\
  6076--6085, 2017.

\bibitem[Simonyan \& Zisserman(2014)Simonyan and Zisserman]{simonyan2014very}
Karen Simonyan and Andrew Zisserman.
\newblock Very deep convolutional networks for large-scale image recognition.
\newblock \emph{arXiv preprint arXiv:1409.1556}, 2014.

\bibitem[Smith(2017)]{smith2017cyclical}
Leslie~N Smith.
\newblock Cyclical learning rates for training neural networks.
\newblock In \emph{Applications of Computer Vision (WACV), 2017 IEEE Winter
  Conference on}, pp.\  464--472. IEEE, 2017.

\bibitem[Smith et~al.(2017)Smith, Kindermans, and Le]{smith2017don}
Samuel~L Smith, Pieter-Jan Kindermans, and Quoc~V Le.
\newblock Don't decay the learning rate, increase the batch size.
\newblock \emph{arXiv preprint arXiv:1711.00489}, 2017.

\bibitem[Vaswani et~al.(2017)Vaswani, Shazeer, Parmar, Uszkoreit, Jones, Gomez,
  Kaiser, and Polosukhin]{vaswani2017attention}
Ashish Vaswani, Noam Shazeer, Niki Parmar, Jakob Uszkoreit, Llion Jones,
  Aidan~N Gomez, {\L}ukasz Kaiser, and Illia Polosukhin.
\newblock Attention is all you need.
\newblock In \emph{Advances in Neural Information Processing Systems}, pp.\
  5998--6008, 2017.

\bibitem[Wilson et~al.(2017)Wilson, Roelofs, Stern, Srebro, and
  Recht]{wilson2017marginal}
Ashia~C Wilson, Rebecca Roelofs, Mitchell Stern, Nati Srebro, and Benjamin
  Recht.
\newblock The marginal value of adaptive gradient methods in machine learning.
\newblock In \emph{Advances in Neural Information Processing Systems}, pp.\
  4151--4161, 2017.

\bibitem[Yosinski et~al.(2014)Yosinski, Clune, Bengio, and
  Lipson]{yosinski2014transferable}
Jason Yosinski, Jeff Clune, Yoshua Bengio, and Hod Lipson.
\newblock How transferable are features in deep neural networks?
\newblock In \emph{Advances in neural information processing systems}, pp.\
  3320--3328, 2014.

\end{thebibliography}
\bibliographystyle{iclr2019_conference}


\newpage
\section*{Appendix}

\begin{figure}
    \centering
    \includegraphics[width=\textwidth]{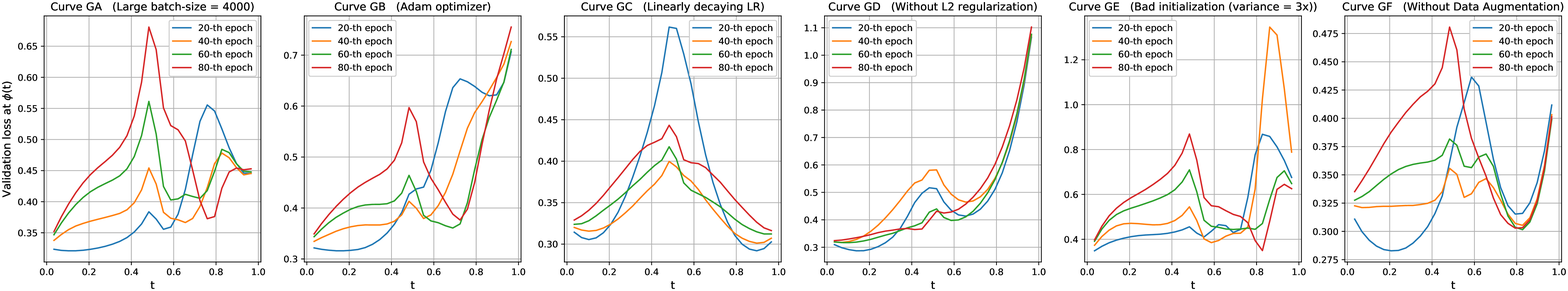}
    \caption{Validation loss corresponding to models on the 6 different curves}
    \label{fig:testlossCF}
\end{figure}
\begin{figure}
    \centering
    \includegraphics[width=\textwidth]{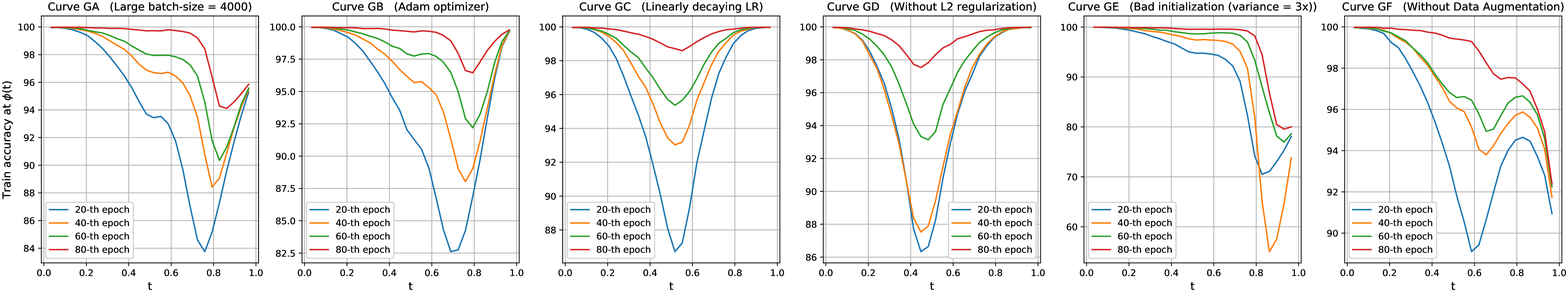}
    \caption{Training accuracy corresponding to models on the 6 different curves}
    \label{fig:trainaccCF}
\end{figure}
\begin{figure}
    \centering
    \includegraphics[width=\textwidth]{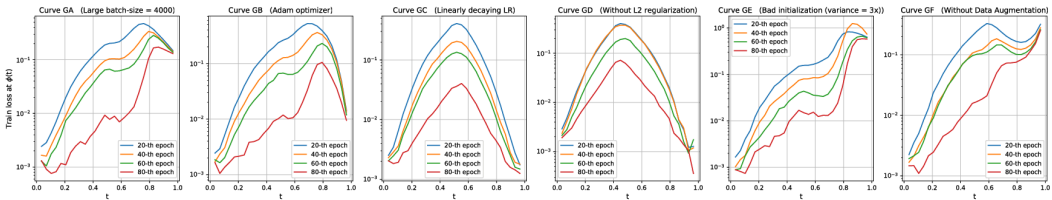}
    \caption{Training loss corresponding to models on the 6 different curves.}
    \label{fig:trainlossCF}
\end{figure}

\section{Additional Results on Robustness of MC}
\label{sec:appendixmc}
\subsection{Training Details}
The learning rate is initialized to $0.05$ and scaled down by a factor of $5$ at epochs $\{60,120,160\}$ (step decay). We use a training batch size of 100, momentum of 0.9, and a weight decay of $0.0005$. Elements of the weight vector corresponding to a neuron are initialized randomly from the normal distribution  $\mathcal{N}(0,\,\sqrt[]{2/n})$ where $n$ is the number of inputs to the neuron. We also use data augmentation by random cropping of input images.

\subsection{Plots}
Figures \ref{fig:testlossCF}, \ref{fig:trainaccCF} and \ref{fig:trainlossCF} show the Validation Loss, Training Accuracy and Training Loss respectively for the curves joining the $6$ pairs discussed in Section \ref{resilience}. These results too, confirm the overfitting or poor generalization tendency of models on the curve.

\subsection{t-SNE visualization for the 7 modes}
\begin{figure}[ht]
\vskip 0.2in
\begin{center}
\centerline{\includegraphics[scale = 0.4]{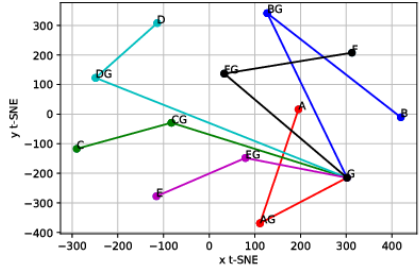}}
\caption{Representing the modes and their connecting point using t-SNE}
\label{constellation}
\end{center}
\vskip -0.2in
\end{figure}

We use t-SNE \citep{maaten2008visualizing} to visualize these $7$ modes and the $\theta$ points that define the connectivity for the $6$ pairs presented in Section \ref{resilience}, in a $2$-dimensional plot in Figure \ref{constellation}. Since t-SNE is known to map only local information correctly and not preserve global distances, we caution the reader about the limited interpretability of this visualization, it is presented simply to establish the notion of connected modes. 

\section{Additional SGDR Results}
\label{sec:appendixsgdr}

\subsection{Additional results}
\begin{figure}
    \centering
    \includegraphics[ width = \textwidth]{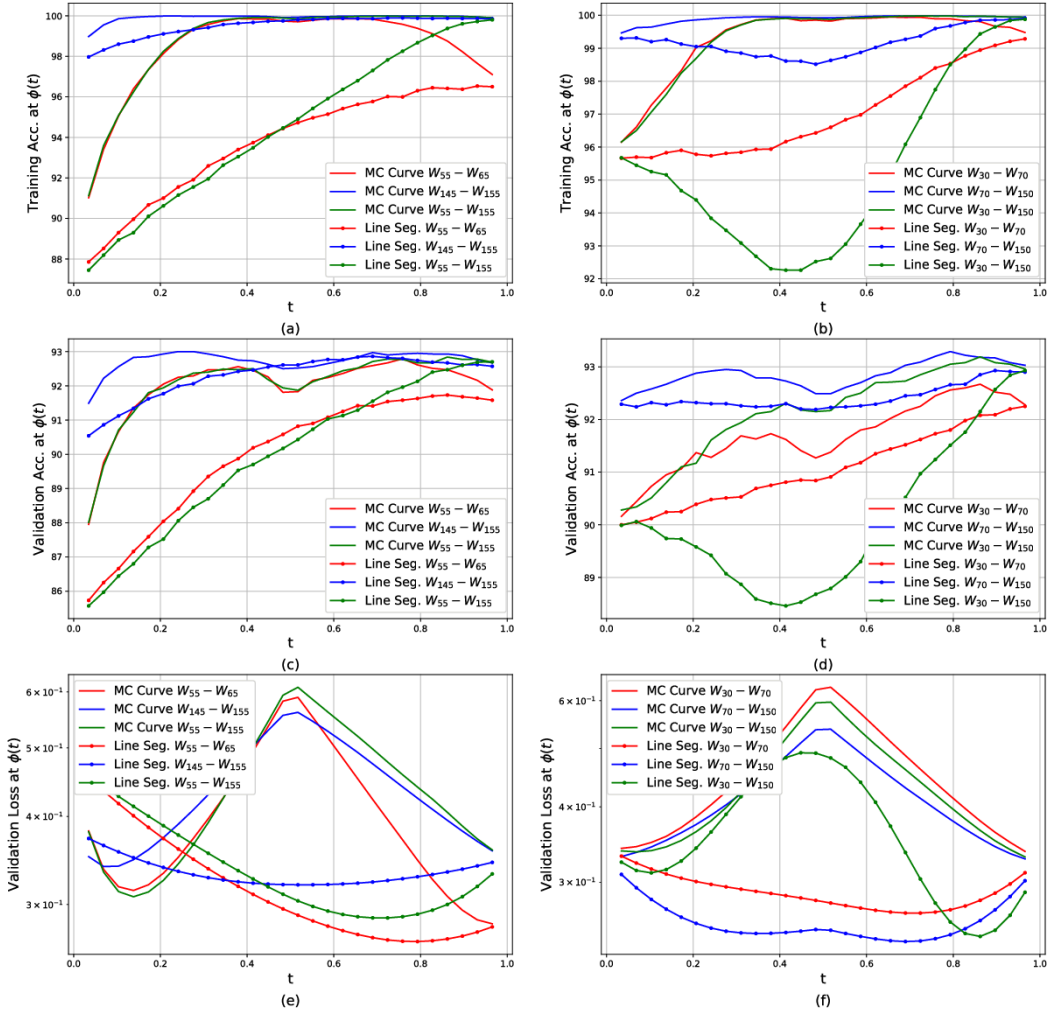}
    \caption{\textbf{Left} Column: Connecting iterates from SGD with step-decay learning rate scheme \textbf{Right} Column: Connecting iterates from SGDR  \textbf{Top} Row: Training Accuracy on the curve found through Mode Connectivity (MC Curve) and on the line segment (Line Seg.) joining iterates from SGDR and SGD. \textbf{Middle} row: Validation Accuracy on the curve found through Mode Connectivity (MC Curve) and on the line segment (Line Seg.) joining iterates from SGDR and SGD.  \textbf{Bottom} row Validation Loss on the curve found through Mode Connectivity (MC Curve) and on the line segment (Line Seg.) joining iterates from SGDR and SGD.}
    \label{fig:addnalsec2plot}
\end{figure}

For completeness, in Figure \ref{fig:addnalsec2plot}, we present the Validation loss, Validation accuracy and Training accuracy results for the curves and line segments joining iterates from SGDR and SGDR discussed in Figure~\ref{fig:sgdr_sgd_modeconn}(c) and (d).

\subsection{Projecting iterates}
\label{section:projection}
The $W_n$ in Figure \ref{fig:sgdr_mc_contours} is equivalent to $$W_n = P_c(w_n) = {\lambda^\star}^\top \begin{bmatrix}
w_{70} \\
w_{150} \\
\theta
\end{bmatrix} $$
\\
$$
\text{where} \; {\lambda^\star} = \text{argmin}_{\lambda \in \mathbb{R}^3} ||\lambda^\top \begin{bmatrix}
           w_{70} \\
           w_{150} \\
           \theta
         \end{bmatrix} - w_n ||_2^2 $$ 
meaning it is the point on the plane (linear combination of $w_{70}, w_{150} \; \text{and} \; \theta$) with the least $l$-2 distance from the original point (iterate in this case). 

\subsection{Connecting modes $w_{30}$ and $w_{70}$ from SGDR}
\label{section:sgdr30to70}
\begin{figure}[ht]
\begin{center}
\centerline{\includegraphics[scale = 0.2]{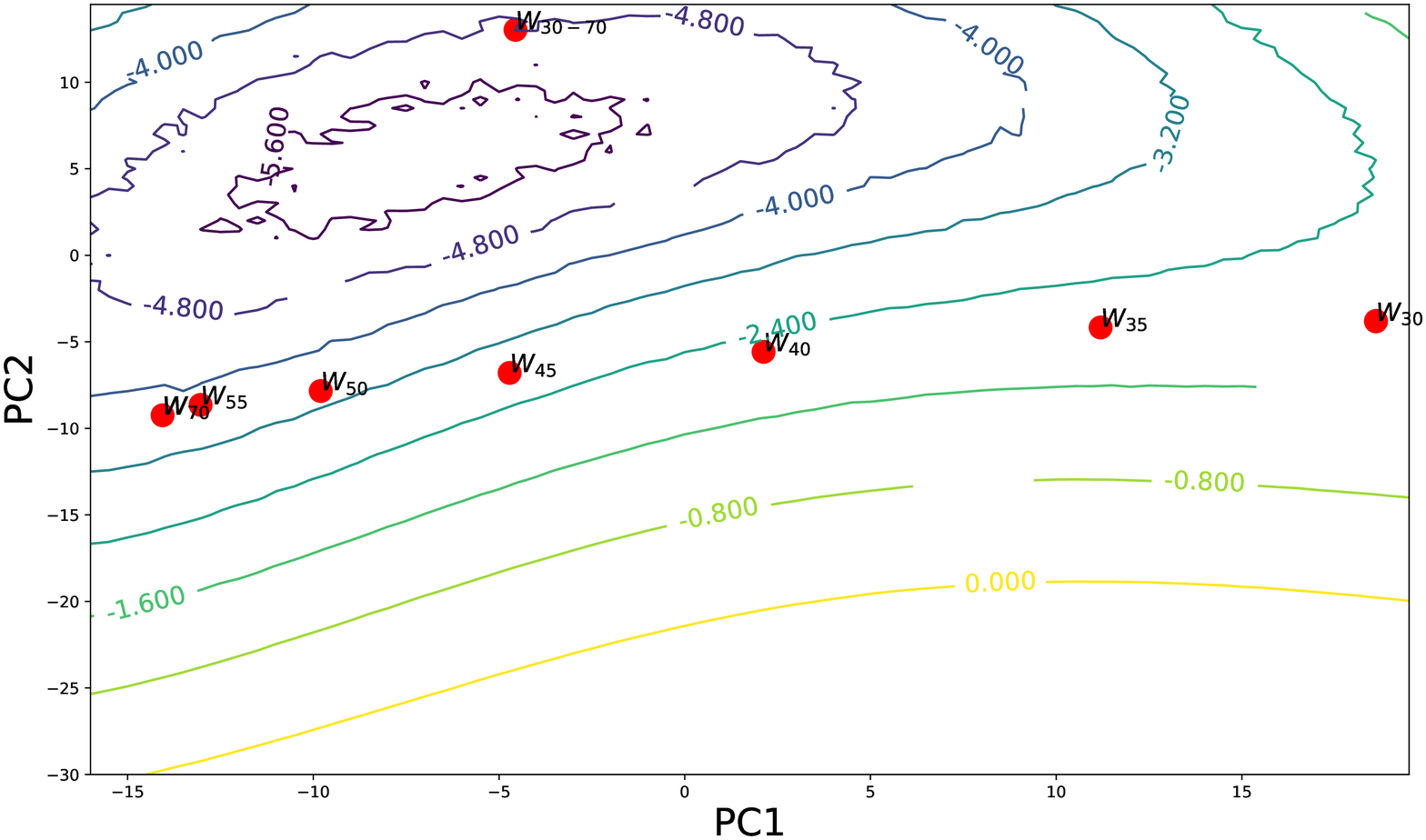}}
\caption{Training Loss Surface (log scale) for points on the plane defined by $\{w_{30},w_{70},w_{30-70}\}$ including projections of iterates on this plane}
\label{app_train_contour}
\end{center}
\vskip -0.2in
\end{figure}

\begin{figure}[ht]
\begin{center}
\centerline{\includegraphics[scale = 0.2]{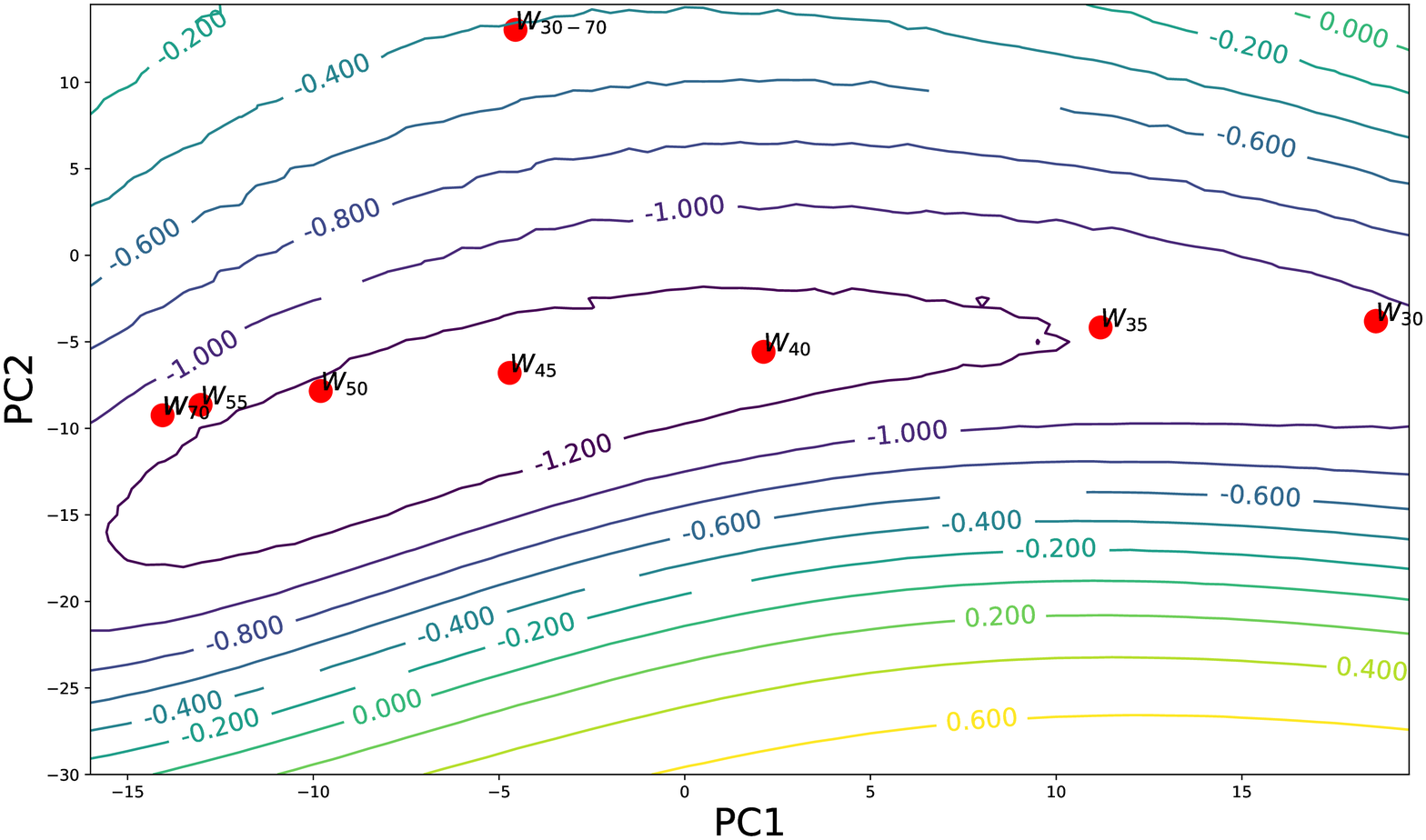}}
\caption{Validation Loss Surface (log scale) for points on the plane defined by $\{w_{30},w_{70},w_{30-70}\}$ including projections of iterates on this plane}
\label{app_val_contour}
\end{center}
\vskip -0.2in
\end{figure}

In Section 3, we present some experiments and make observations on the trajectory of SGDR by using the plane defined by the points $w_{70}$, $w_{150}$ and $w_{70-150}$. Here we plot the training loss surface in Figure \ref{app_train_contour} and the validation loss surface in Figure \ref{app_val_contour} for another plane defined by SGDR's iterates $w_{30}, w_{70}$ and their connection $w_{30-70}$ to ensure the reader that the observations made are general enough.

\section{SGDR CCA Heatmaps}
\label{sec:appendixsgdrsvcca}
\begin{figure}
    \centering
    \includegraphics[width=\textwidth]{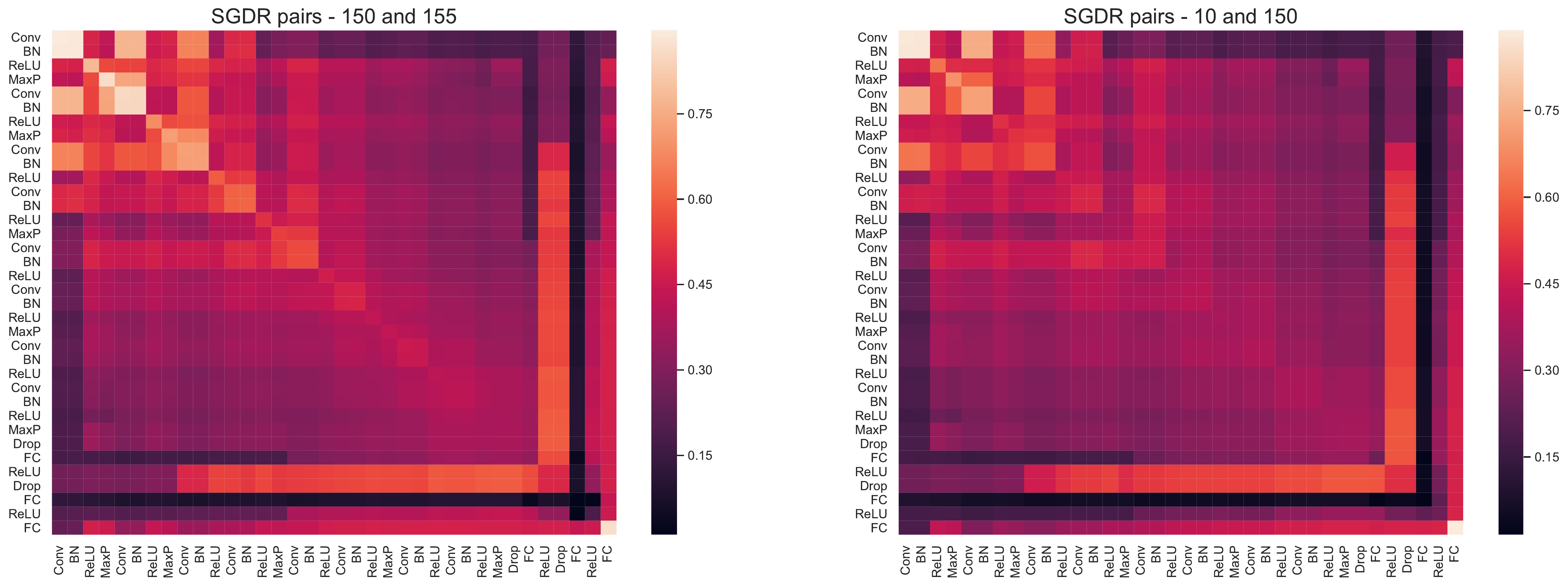}
    \caption{CCA similarity scores between two pairs of models. (a) comparings models at epochs 150 and 155, (b) comparing models at epochs 10 and 150.  The $i,j$-th cell in each pane represents the CCA similarity between layer $i$ of $\textit{w}_a$ (model at epoch a) and layer $j$ of model $\textit{w}_b$ (model at epoch b).}
    \label{fig:SGDR_heatmaps}
\end{figure}

In Figure \ref{fig:SGDR_heatmaps}, we present the CCA similarity plots comparing two pairs of models: epochs 10 and 150, and epochs 150 and 155. The $(i,j)^{th}$ block of the matrix denotes the correlation between the $i^{th}$ layer of the first model and the $j^{th}$ layer of the other. A high correlation implies that the layers learn similar representations and vice versa. We present the former to compare against the typical stepwise or linear decay of SGD, and the latter to demonstrate the immediate effect of restarting on the model. \citet{raghu2017svcca} showed in their work that for typical SGD training, a CCA similarity plot between a partially and completed trained network reveals that the activations of the shallower layers bears closer resemblance in the two models than the deeper layers. We note that, despite the restart, a similar tendency is seen in SGDR training as well. This again suggests that the restart does not greatly impact the model, both in weights and representations, and especially so in the shallower layers. A comparison of epochs 150 and 155, i.e., before and after a restart also stands as evidence for this hypothesis.
\section{Warmup Experiments on ResNet-18 and ResNet-32}
\label{sec:appendixresnetwarmups}
In Figure~\ref{fig:warmup_lr}(d), we show that the stability induced by warmup when training with large batches and learning rates can also be obtained by holding the FC stack frozen. This experiment was conducted on the VGG-11 network \citep{simonyan2014very}. To demonstrate the generality of our claim, we present additional experiments on two ResNet architectures: 18 and 32. The setup for this experiment is identical to the VGG-11 one with one change: instead of the learning rate being set to $2.5$, which is the learning rate for SB ($0.05$) times the batch size increase ($50\times$), we set it to $5.0$ since SB training is better with $0.1$. For the warmup case, we linearly increase the learning rate from $0$ to $5$ again for $20$ epochs. Experiments on other configurations yielded similar results. Whether these results remain true also for training larger datasets, such as ImageNet, remains to be shown and is a topic of future research.

\begin{figure}
    \centering
    \begin{subfigure}[t]{0.45\textwidth}
        \includegraphics[width=\textwidth]{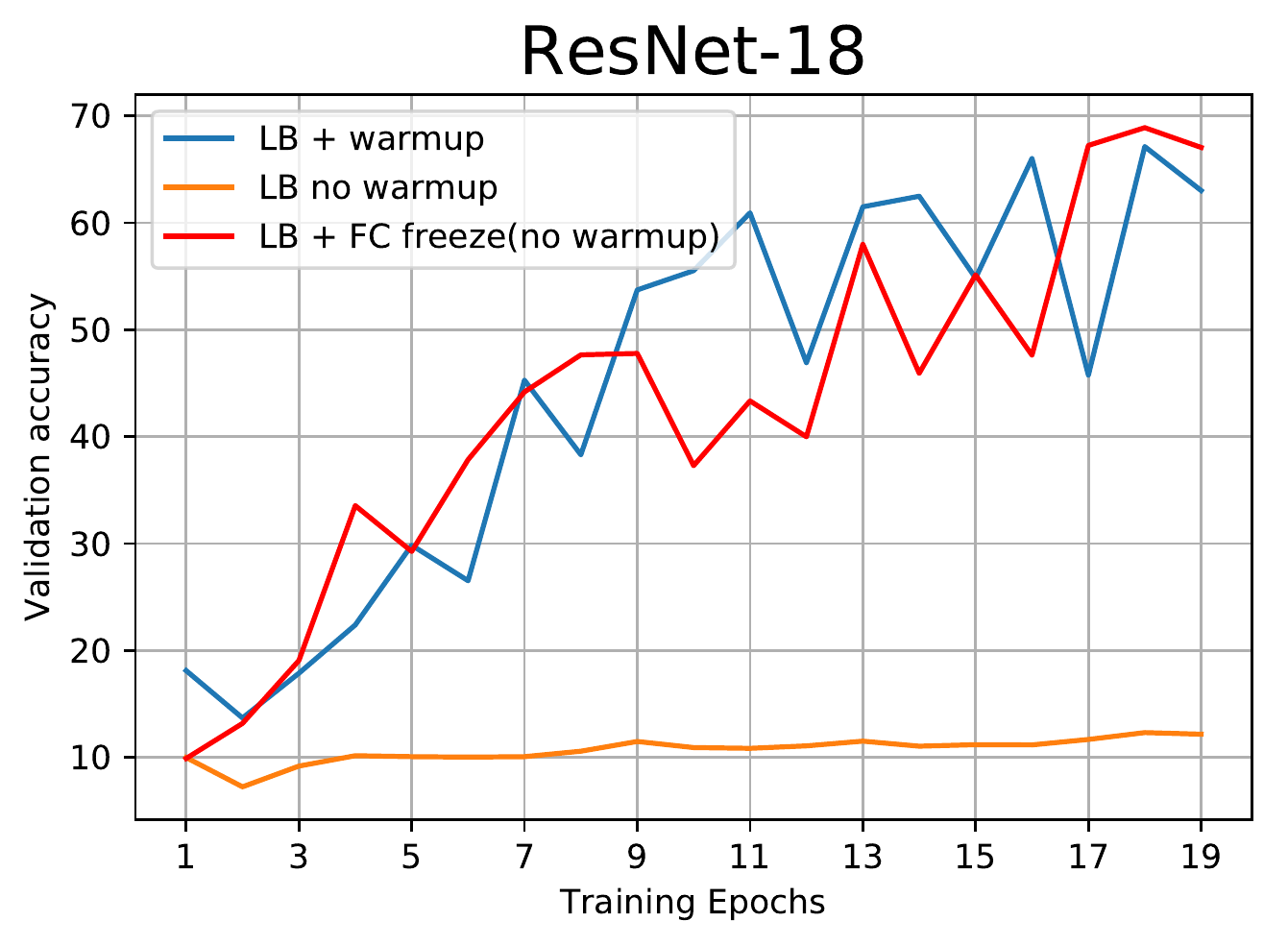}
        \centering
        
    \end{subfigure}
    \quad
    \begin{subfigure}[t]{0.45\textwidth}
        \includegraphics[width=\textwidth]{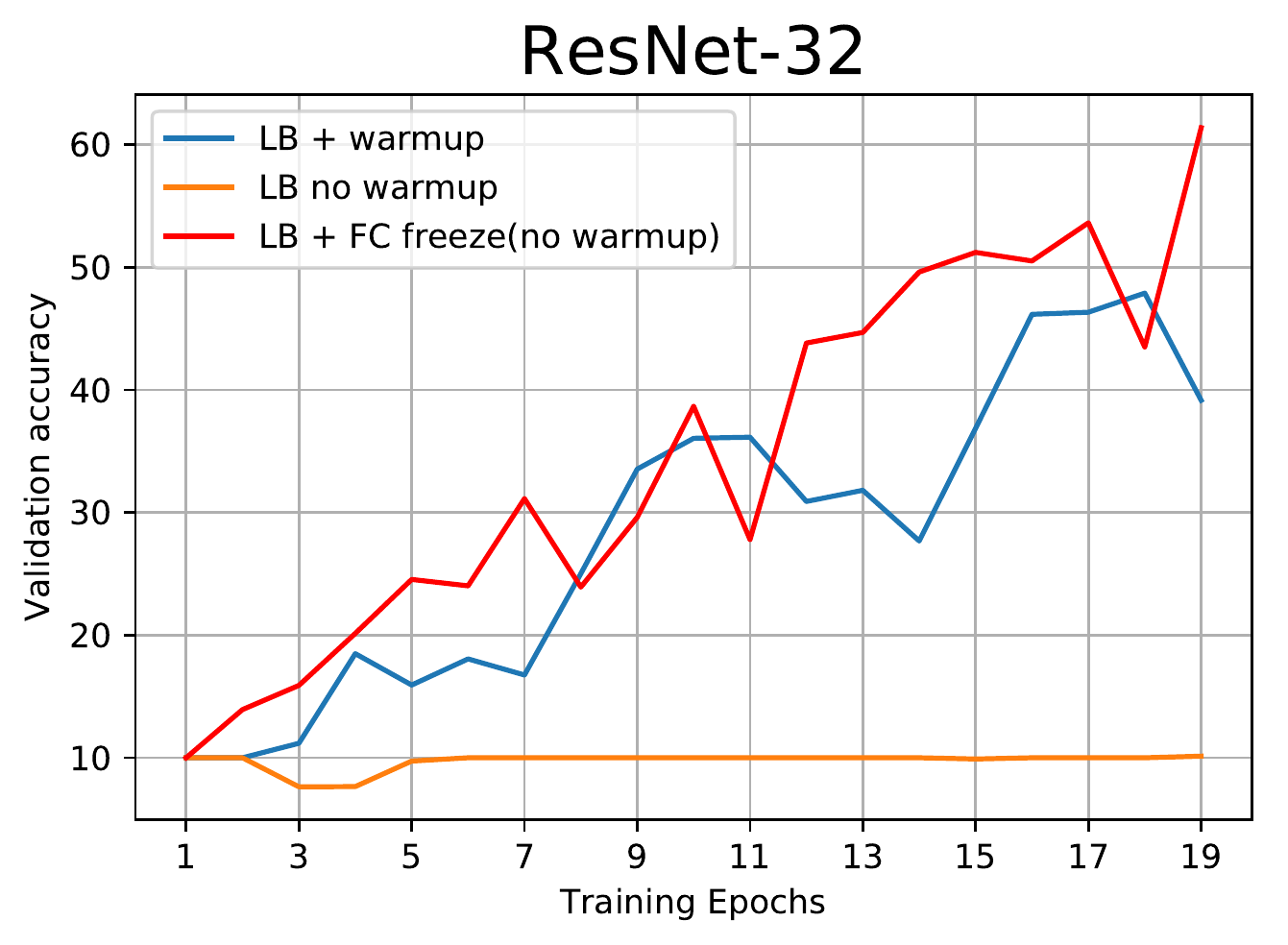}
        \centering
        
    \end{subfigure}
    \\
    \caption{Experiment comparing warmup and FC freezing strategies on ResNet architectures. }
\end{figure}
\end{document}